\documentclass[shortpaper,9pt,twoside,preprint]{ieeecolor_preprint}
\usepackage{cite}
\usepackage{amsmath,amssymb,amsfonts}
\usepackage{algorithmic}
\usepackage{graphicx}
\usepackage{textcomp}
\def\BibTeX{{\rm B\kern-.05em{\sc i\kern-.025em b}\kern-.08em
    T\kern-.1667em\lower.7ex\hbox{E}\kern-.125emX}}

\usepackage{nicefrac}
\usepackage[ruled,noline]{algorithm2e}
\usepackage{booktabs}
\usepackage{array}

\newtheorem{theorem}{Theorem}
\newtheorem{definition}{Definition}

\usepackage[hidelinks]{hyperref}

\newcolumntype{R}[1]{>{\raggedleft\arraybackslash}b{\dimexpr #1\linewidth}<{\hspace*{3pt}}}

\newcommand{\E}{\mathop{\mathbb{E}}} 
\newcommand{\TV}{\operatorname{TV}} 
\newcommand{\KL}{\operatorname{KL}} 
\newcommand{\A}{A^{\pi_k}(s,a)} 
\newcommand{\epsgen}{\epsilon_{\textnormal{GPI}}} 
\newcommand{\deltagen}{\delta_{\textnormal{GPI}}} 
\newcommand{\fess}{f_{\textnormal{ESS}}} 
\newcommand{\ftv}{f_{\textnormal{TV}}} 

\newcommand{\secref}[1]{Section~\ref{#1}}

\newcommand{\thmref}[1]{Theorem~\ref{#1}}
\newcommand{\defref}[1]{Definition~\ref{#1}}

\newcommand{\algref}[1]{Algorithm~\ref{#1}}
\newcommand{\figref}[1]{Fig.~\ref{#1}}
\newcommand{\tabref}[1]{Table~\ref{#1}}

\begin{document}


\title{Generalized Policy Improvement Algorithms\\with Theoretically Supported Sample Reuse}
\author{James Queeney, Ioannis Ch. Paschalidis, \IEEEmembership{Fellow, IEEE}, and Christos G. Cassandras, \IEEEmembership{Life Fellow, IEEE}
\thanks{Accepted for publication in IEEE Transactions on Automatic Control. This research was partially supported by the NSF under grants ECCS-1931600, DMS-1664644, CNS-1645681, IIS-1914792, DEB-2433726, ECCS-2317079, and CCF-2200052, by the ONR under grant N00014-19-1-2571, by the NIH under grants R01 GM135930 and UL54 TR004130, by the DOE under grants DE-AR0001282 and DE-EE0009696, by AFOSR under grant FA9550-19-1-0158, by the MathWorks, and by Boston University. JQ performed the majority of this work while with Boston University. He is currently exclusively supported by Mitsubishi Electric Research Laboratories.}
\thanks{James Queeney is with Mitsubishi Electric Research Laboratories, Cambridge, MA 02139 USA (e-mail: queeney@merl.com).}
\thanks{Ioannis Ch. Paschalidis and Christos G. Cassandras are with the Department of Electrical and Computer Engineering and Division of Systems Engineering, Boston University, Boston, MA 02215 USA (e-mail: yannisp@bu.edu; cgc@bu.edu).}
}

\maketitle


\begin{abstract}
We develop a new class of model-free deep reinforcement learning algorithms for data-driven, learning-based control. Our Generalized Policy Improvement algorithms combine the policy improvement guarantees of on-policy methods with the efficiency of sample reuse, addressing a trade-off between two important deployment requirements for real-world control: (i)~practical performance guarantees and (ii)~data efficiency. We demonstrate the benefits of this new class of algorithms through extensive experimental analysis on a broad range of simulated control tasks.
\end{abstract}




\section{Introduction}

Model-free deep reinforcement learning (RL) represents a successful framework for data-driven control in systems with unknown or complex dynamics \cite{busoniu_2018,recht_2019}, where a learning-based approach to control is necessary. Model-free deep RL algorithms have demonstrated impressive performance on a variety of simulated control tasks in recent years \cite{duan_2016}, and serve as fundamental building blocks in other RL approaches such as model-based~\cite{kurutach_2018,janner_2019,rajeswaran_2020}, offline~\cite{wu_2019,kumar_2020,kidambi_2020}, and safe~\cite{achiam_2017,fisac_2019,wabersich_2022,brunke_2022,paternain_2023} control methods. 

Despite this success, adoption of deep RL methods has remained limited for high-stakes real-world control, where stable performance is critical and data collection can be both expensive and time-consuming. In order to reliably deploy data-driven control methods in these real-world settings, we require algorithms that (i)~provide practical guarantees on performance throughout training and (ii)~make efficient use of data collected in the environment. Unfortunately, these represent competing goals in existing model-free deep RL methods.

Off-policy deep RL algorithms achieve data efficiency through the use of a replay buffer during training, which allows samples to be used for multiple policy updates. Typically, the replay buffer stores millions of samples, which enables state-of-the-art performance on popular benchmarks. However, this aggressive form of sample reuse lacks practical performance guarantees, as both the size of the replay buffer and the sampling method for policy updates are treated as hyperparameters. This prevents off-policy deep RL from being a broadly applicable solution for real-world control, and other data-driven methods are required.

On-policy deep RL algorithms, on the other hand, only use data collected under the current policy in order to provide worst-case performance guarantees at every update. By doing so, on-policy methods guarantee \emph{approximate policy improvement} throughout training, which is often a prerequisite for the deployment of data-driven control in real-world systems. In addition, on-policy approaches require significantly less computation and memory compared to off-policy algorithms \cite{grossman_2023}, making them a viable option in settings that preclude the use of large replay buffers. For these reasons, we consider on-policy algorithms as our starting point in this work. However, the requirement of on-policy data results in high sample complexity and slow learning, which limits the effectiveness of these algorithms in practice.

Given our goal of reliable and efficient training, in this work we combine the policy improvement benefits of on-policy methods with the efficiency of sample reuse. This paper extends our prior work in \cite{queeney_2021_geppo}, where we generalized the popular on-policy algorithm Proximal Policy Optimization (PPO) \cite{schulman_2017} to incorporate sample reuse. Compared to \cite{queeney_2021_geppo}, our main contributions are as follows:
\begin{enumerate}
\item In \secref{sec:gen_pilb}, we introduce a generalized trust region update that serves as the foundation for a unified class of \emph{Generalized Policy Improvement (GPI)} algorithms. Our GPI framework extends many popular on-policy methods to incorporate sample reuse.
\item In \secref{sec:reuse}, we demonstrate how to optimally reuse data from all recent policies. Our \emph{theoretically supported sample reuse} improves the trade-off between batch size and policy update size throughout training compared to on-policy methods, while retaining the same approximate policy improvement guarantees. This is in contrast to the aggressive sample reuse in off-policy methods that lacks practical performance guarantees.
\item In \secref{sec:gpi}, we introduce three examples of our GPI framework. In addition to the generalized version of PPO from \cite{queeney_2021_geppo}, we also develop generalized versions of the popular on-policy algorithms Trust Region Policy Optimization (TRPO)~\cite{schulman_2015} and On-Policy Maximum a Posteriori Policy Optimization (VMPO)~\cite{song_2020}.
\item In \secref{sec:experiments}, we demonstrate the practical benefits of our GPI algorithms through extensive experimental analysis on a broad range of continuous control benchmarking tasks from the DeepMind Control Suite \cite{tunyasuvunakool_2020}.
\end{enumerate}


\section{Related Work}


\subsection{On-Policy Policy Improvement Methods}

Our work focuses on policy improvement methods, an algorithmic framework that was first introduced by \cite{kakade_2002} in Conservative Policy Iteration. The policy improvement bounds proposed by \cite{kakade_2002} were later refined by \cite{schulman_2015} and \cite{achiam_2017}, making them compatible with the deep RL setting. These advances motivated the development of many popular on-policy policy improvement methods, including TRPO \cite{schulman_2015} and PPO \cite{schulman_2017}. The strong empirical performance of TRPO and PPO has been studied in detail \cite{henderson_2018,engstrom_2020,andrychowicz_2021}, and these algorithms have served as the foundation for many on-policy extensions \cite{achiam_2017,queeney_2021_uatrpo,wang_y_2019,wang_y_2020,cheng_2022}. Given the popularity of TRPO and PPO, we consider both of these algorithms when developing our family of GPI algorithms.

In recent years, on-policy methods have also considered non-parametric target policies obtained by solving policy optimization problems similar to those used in TRPO and PPO \cite{vuong_2018,song_2020}. These target policies are then projected back onto the space of parameterized policies. VMPO \cite{song_2020} considers a target policy induced by a reverse Kullback-Leibler (KL) divergence constraint, and its practical trust region implementation can be interpreted from a policy improvement perspective. Therefore, we also propose a generalized version of VMPO as a representative instance of policy improvement algorithms based on non-parametric target policies.


\subsection{Sample Efficiency with Off-Policy Data}

The main drawback of on-policy policy improvement algorithms is their requirement of on-policy data, which can lead to slow and inefficient learning. Popular off-policy deep RL algorithms~\cite{lillicrap_2016,fujimoto_2018,haarnoja_2018,abdolmaleki_2018} address this issue by storing data in a large replay buffer and aggressively reusing samples for many policy updates. Typically, these algorithms do not explicitly control the bias introduced by off-policy data, and as a result do not provide practical policy improvement guarantees. 

Methods have been proposed to address the bias introduced by off-policy data. One line of work combines on-policy and off-policy updates to address this concern \cite{odonoghue_2017,gu_2017_qprop,gu_2017_ipg,meng_2022,fakoor_2020,wang_z_2017}, and some of these methods consider regularization terms that are similar to the generalized KL divergence trust regions in our Generalized TRPO and VMPO algorithms. Other approaches control the bias from off-policy data by modifying the use of the replay buffer, such as ignoring samples whose actions are not likely under the current policy \cite{novati_2019} or emphasizing more recent experience \cite{wang_c_2020}. The theoretically supported sample reuse that we consider in our GPI algorithms can be viewed as restricting the size of the replay buffer to only contain recent policies, as well as applying a non-uniform weighting to the samples.


\section{Preliminaries}


\begin{algorithm}[t]
\caption{On-Policy Policy Improvement Algorithms}\label{alg:on}
\KwIn{initial policy $\pi_0$; TV distance trust region parameter $\epsilon$; batch size $N$.}
\BlankLine
\For{$k=0,1,2,\ldots$}{
	\BlankLine
        Collect $N$ samples via trajectory rollouts $\tau \sim \pi_k$.	
	\BlankLine
	Use $N$ samples from $\pi_k$ to approximate the expectations in \eqref{eq:on_trust_tv}. 
	\BlankLine	
	Update policy by approximately solving the optimization problem in \eqref{eq:on_trust_tv}. Implementation varies by algorithm.
	\BlankLine
}
\end{algorithm}



\subsection{Reinforcement Learning Framework}

We represent the sequential decision making problem as an infinite-horizon, discounted Markov decision process (MDP) defined by the tuple $(\mathcal{S}, \mathcal{A}, p, r, \rho_0, \gamma)$, where $\mathcal{S}$ is the set of states, $\mathcal{A}$ is the set of actions, $p: \mathcal{S} \times \mathcal{A} \rightarrow P(\mathcal{S})$ is the transition probability function where $P(\mathcal{S})$ represents the space of probability measures over $\mathcal{S}$, $r: \mathcal{S} \times \mathcal{A} \rightarrow \mathbb{R}$ is the reward function, $\rho_0 \in P(\mathcal{S})$ is the initial state distribution, and $\gamma$ is the discount rate.

We model the agent's decisions as a stationary policy $\pi_{\boldsymbol{\theta}}: \mathcal{S} \rightarrow P(\mathcal{A})$ parameterized by $\boldsymbol{\theta} \in \boldsymbol{\Theta}$. In this work, we consider policies parameterized by neural networks where $\pi_{\boldsymbol{\theta}}(\, \cdot \mid s)$ shares the same support for all $\boldsymbol{\theta} \in \boldsymbol{\Theta}$. This is true of most popular policy parameterizations in deep RL, including the multivariate Gaussian policy we consider in our experiments. In the remainder of the paper, we write $\pi$ without the subscript for ease of notation.

Our goal is to find a policy $\pi$ that maximizes the expected total discounted return $J(\pi) = \E_{\tau \sim \pi} \left[ \sum_{t=0}^{\infty} \gamma^t r(s_t,a_t) \right]$, where $\tau \sim \pi$ represents a trajectory sampled according to $s_0 \sim \rho_0$, $a_t \sim \pi(\, \cdot \mid s_t)$, and $s_{t+1} \sim p(\, \cdot \mid s_t, a_t)$. We denote the state value function of $\pi$ as $V^{\pi}(s) = \E_{\tau \sim \pi} \left[ \sum_{t=0}^{\infty} \gamma^t r(s_t,a_t) \mid s_0 = s \right]$, the state-action value function as $Q^{\pi}(s,a) = \E_{\tau \sim \pi} \left[ \sum_{t=0}^{\infty} \gamma^t r(s_t,a_t) \mid s_0 = s, a_0 = a \right]$, and the advantage function as $A^{\pi}(s,a) = Q^{\pi}(s,a) - V^{\pi}(s)$. A policy $\pi$ also induces a normalized discounted state visitation distribution $d^{\pi}$, where $d^{\pi}(s) = (1-\gamma) \sum_{t=0}^{\infty} \gamma^t \mathbb{P}(s_t = s \mid \rho_0, \pi, p)$. We write the corresponding normalized discounted state-action visitation distribution as $d^{\pi}(s,a) = d^{\pi}(s) \pi(a \mid s)$, where we make it clear from the context whether $d^{\pi}$ refers to a distribution over states or state-action pairs.


\subsection{On-Policy Policy Improvement Algorithms}

Many popular on-policy algorithms can be interpreted as approximately maximizing the following policy improvement lower bound, which was first developed by \cite{kakade_2002} and later refined by \cite{schulman_2015} and \cite{achiam_2017}.

\begin{theorem}[From \cite{achiam_2017}]\label{thm:on_pilb_tv}
Consider any policy $\pi$ and a current policy $\pi_k$. Then, we have that
\begin{multline}\label{eq:on_pilb_tv}
J(\pi) - J(\pi_k) \geq \frac{1}{1-\gamma} \E_{(s,a) \sim d^{\pi_k}} \left[ \frac{\pi(a \mid s)}{\pi_k(a \mid s)} \A \right] \\ - \frac{2 \gamma C^{\pi,\pi_k}}{(1-\gamma)^2} \E_{s \sim d^{\pi_k}} \left[ \TV \left( \pi, \pi_k \right)(s) \right],
\end{multline}
where $\TV \left( \pi, \pi_k \right)(s) = \frac{1}{2} \int_{\mathcal{A}} \left| \pi(a \mid s) - \pi_k(a \mid s) \right| \textnormal{d}a$ represents the Total Variation (TV) distance between the distributions $\pi(\, \cdot \mid s)$ and $\pi_k(\, \cdot \mid s)$, and $C^{\pi,\pi_k} = \max_{s \in \mathcal{S}} \left| \E_{a \sim \pi(\, \cdot \, \mid s)} \left[ \A \right] \right|$.
\end{theorem}

We refer to the first term on the right-hand side of \eqref{eq:on_pilb_tv} as the surrogate objective and the second term as the penalty term. Rather than directly maximize the lower bound in \thmref{thm:on_pilb_tv}, on-policy policy improvement algorithms typically maximize the surrogate objective while bounding the risk of each policy update via a constraint on the penalty term. This leads to updates with the following form.
\begin{definition}\label{def:on_update}
For a given choice of trust region parameter $\epsilon > 0$, the \emph{on-policy trust region update} has the form
\begin{equation}\label{eq:on_trust_tv}
\begin{split}
\pi_{k+1} = \arg \max_{\pi} \,\, & \E_{(s,a) \sim d^{\pi_k}} \left[ \frac{\pi(a \mid s)}{\pi_k(a \mid s)} \A \right] \\
\textnormal{s.t.} \,\, & \E_{s \sim d^{\pi_k}} \left[ \TV \left( \pi, \pi_k \right)(s) \right] \leq \frac{\epsilon}{2}.
\end{split}
\end{equation}
\vspace{0.0em}
\end{definition}

By maximizing the surrogate objective from \eqref{eq:on_pilb_tv} while constraining the magnitude of the penalty term from \eqref{eq:on_pilb_tv}, the trust region update in \eqref{eq:on_trust_tv} limits the \emph{worst-case performance decline} at every update to
\begin{equation}\label{eq:pd_worstcase}
J(\pi_{k+1}) - J(\pi_k) \geq - \frac{\epsilon \gamma C^{\pi_{k+1},\pi_k}}{(1-\gamma)^2}.
\end{equation}
Therefore, we say that on-policy algorithms based on the trust region update in \eqref{eq:on_trust_tv} deliver \emph{approximate} policy improvement guarantees. In addition, practical deep RL implementations of this update introduce additional approximations through the use of sample-based estimates.

The high-level framework of on-policy policy improvement algorithms is described in \algref{alg:on}. The difference between popular on-policy algorithms is primarily due to how they approximately solve the optimization problem in \eqref{eq:on_trust_tv}. We provide details for PPO, TRPO, and VMPO in \secref{sec:gpi}.


\section{Generalized Trust Region Update}\label{sec:gen_pilb}

The expectations that appear in the surrogate objective and penalty term of \eqref{eq:on_pilb_tv} can be estimated using samples collected under the current policy $\pi_k$. Therefore, this bound motivates algorithms that are practical to implement, but may result in slow learning due to their requirement of on-policy data. The goal of our work is to improve the efficiency of on-policy algorithms through sample reuse, without sacrificing their approximate policy improvement guarantees. In order to relax the on-policy requirement of the expectations that appear in \eqref{eq:on_pilb_tv}, we leverage our Generalized Policy Improvement lower bound from \cite{queeney_2021_geppo}.

\begin{theorem}[From \cite{queeney_2021_geppo}]\label{thm:gen_pilb_tv}
Consider any policy $\pi$ and $M$ prior policies $\pi_{k-i}$, $i=0,\ldots,M-1$, where $k \geq M-1$. Let $\nu$ be any choice of mixture distribution over these $M$ prior policies, where $0 \leq \nu_i \leq 1$ is the probability assigned to $\pi_{k-i}$, $\sum_{i=0}^{M-1} \nu_i = 1$, and $\E_{i \sim \nu} \left[ \, \cdot \, \right]$ represents an expectation determined by this mixture distribution. Then, we have that
\begin{multline}\label{eq:gen_pilb_tv}
J(\pi) - J(\pi_k) \\ \geq  \frac{1}{1-\gamma} \E_{i \sim \nu} \left[ \E_{(s,a) \sim d^{\pi_{k-i}}} \left[ \frac{\pi(a \mid s)}{\pi_{k-i}(a \mid s)} \A \right] \right] \\ - \frac{2 \gamma C^{\pi,\pi_k}}{(1-\gamma)^2} \E_{i \sim \nu} \left[ \E_{s \sim d^{\pi_{k-i}}} \left[ \TV \left( \pi, \pi_{k-i} \right)(s) \right] \right],
\end{multline}
where $C^{\pi,\pi_k}$ is defined as in \thmref{thm:on_pilb_tv}.
\end{theorem}

The expectations that appear in our Generalized Policy Improvement lower bound can be estimated using a mixture of samples collected under prior policies, so \thmref{thm:gen_pilb_tv} provides insight into how we can reuse samples while still providing guarantees on performance throughout training. The cost of sample reuse is that the penalty term now depends on the expected TV distance between the new policy and our prior policies, rather than the current policy. Note that we recover the on-policy lower bound when $\nu$ is chosen to place all weight on the current policy, so \thmref{thm:on_pilb_tv} is a special case of \thmref{thm:gen_pilb_tv}.

Because the structure of our generalized lower bound remains the same as the on-policy lower bound, we can use the same techniques to motivate practical policy improvement algorithms. Just as the on-policy lower bound motivated the policy update in \eqref{eq:on_trust_tv}, we can use \thmref{thm:gen_pilb_tv} to motivate a generalized policy update of the form
\begin{align}
\pi_{k+1} = \arg \max_{\pi} \,\, & \E_{i \sim \nu} \left[ \E_{(s,a) \sim d^{\pi_{k-i}}} \left[ \frac{\pi(a \mid s)}{\pi_{k-i}(a \mid s)} \A \right] \right] \nonumber \\
\textnormal{s.t.} \,\, & \E_{i \sim \nu} \left[ \E_{s \sim d^{\pi_{k-i}}} \left[ \TV \left( \pi, \pi_{k-i} \right)(s) \right] \right] \leq \frac{\epsilon}{2}, \label{eq:gen_trust_tv_full}
\end{align}
where $\epsilon$ represents the same trust region parameter used in the on-policy case. By the triangle inequality of TV distance, we have that 
\begin{multline}\label{eq:tv_triangle}
\E_{i \sim \nu} \left[ \E_{s \sim d^{\pi_{k-i}}} \left[ \TV \left( \pi, \pi_{k-i} \right)(s) \right] \right] \\ \leq \E_{i \sim \nu} \left[ \E_{s \sim d^{\pi_{k-i}}} \left[ \TV \left( \pi, \pi_{k} \right)(s) \right] \right] \\+ \E_{i \sim \nu} \left[ \E_{s \sim d^{\pi_{k-i}}} \left[ \TV \left( \pi_{k}, \pi_{k-i} \right)(s) \right] \right], 
\end{multline}
where the first term on the right-hand side is the expected one-step TV distance and the second term does not depend on $\pi$. Therefore, we can satisfy the trust region in \eqref{eq:gen_trust_tv_full} by controlling the expected one-step TV distance, which is often easier to work with in practice. For an appropriate choice of $\epsgen$, this leads to the following policy update.
\begin{definition}\label{def:gen_update}
For a given choice of trust region parameter $\epsgen > 0$, the \emph{generalized trust region update} has the form
\begin{align}
\pi_{k+1} = \arg \max_{\pi} \,\, & \E_{i \sim \nu} \left[ \E_{(s,a) \sim d^{\pi_{k-i}}} \left[ \frac{\pi(a \mid s)}{\pi_{k-i}(a \mid s)} \A \right] \right] \nonumber \\
\textnormal{s.t.} \,\, & \E_{i \sim \nu} \left[ \E_{s \sim d^{\pi_{k-i}}} \left[ \TV \left( \pi, \pi_{k} \right)(s) \right] \right] \leq \frac{\epsgen}{2}. \label{eq:gen_trust_tv_one}
\end{align}
\phantom{}
\end{definition}

Similar to the on-policy trust region update in \eqref{eq:on_trust_tv}, the generalized trust region update in \eqref{eq:gen_trust_tv_one} also provides approximate policy improvement guarantees due to its connection to the Generalized Policy Improvement lower bound in \thmref{thm:gen_pilb_tv}. In order to deliver these approximate policy improvement guarantees, the generalized update still depends on the advantage function with respect to the \emph{current} policy $\pi_k$, which must be approximated using off-policy estimation techniques in practice.

Next, we describe how to select the generalized trust region parameter $\epsgen$ and mixture distribution $\nu$ over prior policies in order to provide guarantees on the risk of every policy update while optimizing key quantities of interest. Principled choices of $\epsgen$ and $\nu$ result in \emph{theoretically supported sample reuse}.


\section{Theoretically Supported Sample Reuse}\label{sec:reuse}


\subsection{Generalized Trust Region Parameter}

From \eqref{eq:tv_triangle}, we see that the generalized update in \defref{def:gen_update} satisfies the trust region in \eqref{eq:gen_trust_tv_full} for any $\epsgen$ such that
\begin{equation}\label{eq:epsgen_adapt}
\frac{\epsgen}{2} \leq \frac{\epsilon}{2} - \E_{i \sim \nu} \left[ \E_{s \sim d^{\pi_{k-i}}} \left[ \TV \left( \pi_{k}, \pi_{k-i} \right)(s) \right] \right]. 
\end{equation}
While the adaptive choice of $\epsgen$ given by \eqref{eq:epsgen_adapt} will successfully control the magnitude of the penalty term in \eqref{eq:gen_pilb_tv}, it only indirectly provides insight into how $\epsgen$ depends on the choice of $\nu$. In order to establish a direct connection between $\epsgen$ and $\nu$, our analysis considers a slightly stronger trust region update given by
\begin{align}
\pi_{k+1} = \arg \max_{\pi} \,\, & \E_{i \sim \nu} \left[ \E_{(s,a) \sim d^{\pi_{k-i}}} \left[ \frac{\pi(a \mid s)}{\pi_{k-i}(a \mid s)} \A \right] \right] \nonumber \\
\textnormal{s.t.} \,\, & \E_{s \sim d^{\pi_{k-i}}} \left[ \TV \left( \pi, \pi_{k} \right)(s) \right] \leq \frac{\epsgen}{2}, \nonumber \\ 
&\qquad \qquad i=0,\ldots,M-1. \label{eq:gen_trust_tv_one_strong}
\end{align}

\begin{theorem}\label{thm:epsgen}
Consider policy updates determined by \eqref{eq:gen_trust_tv_one_strong}, where
\begin{equation}\label{eq:epsgen}
\epsgen = \frac{\epsilon}{\E_{i \sim \nu} \left[ \, i + 1 \, \right]}.
\end{equation}
Then, we have that $\pi_{k+1}$ satisfies
\begin{equation}\label{eq:epsgen_bound}
\E_{i \sim \nu} \left[ \E_{s \sim d^{\pi_{k-i}}} \left[ \TV \left( \pi_{k+1}, \pi_{k-i} \right)(s) \right] \right] \leq \frac{\epsilon}{2},
\end{equation}
so that $\pi_{k+1}$ is a feasible solution to \eqref{eq:gen_trust_tv_full}.
\end{theorem}

\begin{proof}
Using the triangle inequality for TV distance, we have
\begin{equation*}
\begin{split}
&\E_{i \sim \nu} \left[ \E_{s \sim d^{\pi_{k-i}}} \left[ \TV \left( \pi_{k+1}, \pi_{k-i} \right)(s) \right] \right] \\ 
&\qquad \leq \E_{i \sim \nu} \left[ \sum_{j=0}^i \E_{s \sim d^{\pi_{k-i}}} \left[ \TV \left( \pi_{k-j+1}, \pi_{k-j} \right)(s) \right] \right] \\
&\qquad \leq \E_{i \sim \nu} \left[ \frac{\epsgen}{2} \cdot \left( i + 1 \right) \right] = \frac{\epsgen}{2} \cdot \E_{i \sim \nu} \left[ \, i + 1 \, \right] = \frac{\epsilon}{2},
\end{split}
\end{equation*}
where we have used the trust region constraints in \eqref{eq:gen_trust_tv_one_strong} from all prior updates to bound each term inside the summation by $\nicefrac{\epsgen}{2}$.
\end{proof}

Importantly, \thmref{thm:epsgen} implies that the policy update in \eqref{eq:gen_trust_tv_one_strong} has the same worst-case performance guarantee given by \eqref{eq:pd_worstcase} as the on-policy update in \eqref{eq:on_trust_tv}. In practice, we consider the generalized trust region update in \eqref{eq:gen_trust_tv_one}, which relaxes the trust region in \eqref{eq:gen_trust_tv_one_strong}. We find that the generalized update in \eqref{eq:gen_trust_tv_one} also satisfies \eqref{eq:epsgen_bound} in practice, and tends to be conservative because $\epsgen$ in \eqref{eq:epsgen} is based on applying the triangle inequality between every prior policy. In addition, practical implementations based on clipping mechanisms or backtracking line searches can ensure that \eqref{eq:epsgen_bound} holds. 

The form of $\epsgen$ in \eqref{eq:epsgen} clearly demonstrates the trade-off of sample reuse. As we reuse older data, we must consider smaller one-step trust regions at every policy update in order to guarantee the same level of risk.


\subsection{Mixture Distribution}


\begin{figure}
\centering
\includegraphics[width=1.00\linewidth]{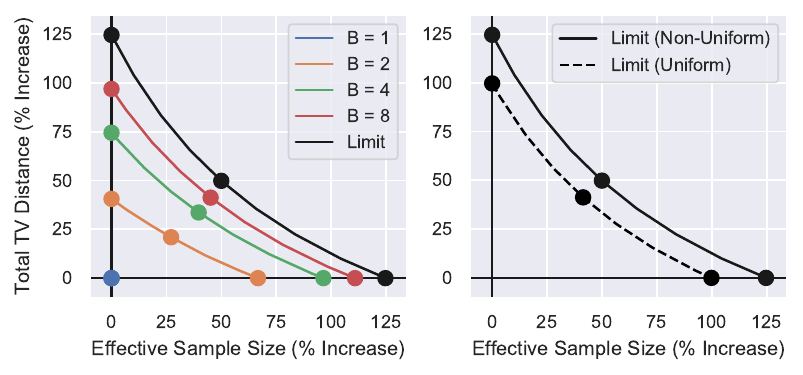}
\caption{Benefit of GPI update compared to on-policy case. Represents percent increases in effective sample size and total TV distance update size for all possible values of $\kappa \in [0,1]$. Markers indicate $\kappa=0.0,0.5,1.0$. Left: Comparison across several values of $B$. Right: Comparison of non-uniform and uniform weights for large $B$.}\label{fig:gpi_nu}
\end{figure}


Despite the need for smaller one-step trust regions at every policy update, we can show that our generalized policy update improves key quantities of interest when the mixture distribution $\nu$ is chosen in a principled manner. Note that $\nu_i > 0$ indicates that data from $\pi_{k-i}$ will be used during updates, so the choice of $\nu$ determines how many prior policies to consider in addition to how to weight their contributions.

We write the on-policy sample size as $N  = Bn$, where $n$ represents the smallest possible batch size for data collection (e.g., the length of one full trajectory) and $B$ is a positive integer. On-policy policy improvement algorithms update the policy according to \eqref{eq:on_trust_tv} after every $N$ samples collected, where the expectations are approximated using empirical averages calculated with these $N$ samples. For the generalized case, we can combine data across several prior policies to construct the batch used for calculating the generalized policy update in \eqref{eq:gen_trust_tv_one}. Therefore, sample reuse allows us to make policy updates after every $n$ samples collected.

It is common to consider auxiliary metrics to inform sampling schemes in deep RL \cite{schaul_2016,debruin_2018}. In order to determine $\nu$, we focus on two important quantities in policy optimization: (i)~effective sample size and (ii)~total TV distance update size. We now define these metrics for our generalized policy updates.

\begin{definition}
The \emph{effective sample size} used for generalized policy updates is given by
\begin{equation*}
\fess(\nu) = \frac{n}{\sum_{i=0}^{M-1} \nu_i^2},
\end{equation*}
where $\nu_i$ represents the probability assigned to data collected under $\pi_{k-i}$, compared to $N=Bn$ in on-policy algorithms.
\end{definition}

\begin{definition}
The \emph{total TV distance update size} of generalized policy updates for every $N$ samples collected is given by
\begin{equation*}
\ftv(\nu) = B \cdot \frac{\epsgen}{2} = \frac{B}{\sum_{i=0}^{M-1} \nu_i \left( i + 1 \right)} \cdot \frac{\epsilon}{2},
\end{equation*}
compared to $\nicefrac{\epsilon}{2}$ in on-policy algorithms.
\end{definition}

The effective sample size \cite{kong_1992} adjusts the sample size to account for increased variance due to non-uniform weights, and reduces to the standard sample size definition of $Mn$ for the case of uniform weights over the last $M$ policies. A larger effective sample size results in more accurate estimates of the expectations that we must approximate in policy updates, and leads to a more diverse batch of data which can be useful when reward signals are sparse \cite{hong_2018}. A larger total TV distance update size, on the other hand, allows for more aggressive exploitation of the available information, which can lead to faster learning throughout training. Although optimizing these metrics does not guarantee improved performance compared to on-policy algorithms, we see in \secref{sec:experiments} that experimentally this is often the case.

Using the following result, we select $\nu$ to optimize $\fess(\nu)$ and $\ftv(\nu)$ relative to the on-policy case.

\begin{theorem}\label{thm:mix}
Fix the trade-off parameter $\kappa \in [0,1]$, and select the mixture distribution $\nu$ according to the convex optimization problem
\begin{align}
\nu^*(\kappa) = &\arg \min_{\nu} \,\, \kappa \cdot \frac{\sum_{i=0}^{\bar{M}-1} \nu_i^2}{c_{\textnormal{ESS}}}  + (1-\kappa) \cdot \frac{\sum_{i=0}^{\bar{M}-1} \nu_i \left( i + 1 \right)}{c_{\textnormal{TV}}}  \nonumber \\
&\quad \,\, \textnormal{s.t.} \quad {\textstyle \sum_{i=0}^{\bar{M}-1} \nu_i^2 } \leq \frac{1}{B}, \quad {\textstyle \sum_{i=0}^{\bar{M}-1} \nu_i \left( i + 1 \right) } \leq B, \nonumber \\
 & \qquad \quad \,\, {\textstyle \sum_{i=0}^{\bar{M}-1} \nu_i } = 1, \,\, \nu_i \geq 0, \,\, i=0,\ldots,\bar{M}-1, \label{eq:mix}
\end{align}
where $c_{\textnormal{ESS}}, c_{\textnormal{TV}} \geq 0$ are scaling coefficients and $\bar{M}$ is large. Then, we have that $\fess(\nu^*) \geq Bn$ and $\ftv(\nu^*) \geq \nicefrac{\epsilon}{2}$.
\end{theorem}

\begin{proof}
Note that $\fess(\nu)$ and $\ftv(\nu)$ only depend on $\nu$ in their denominators. Therefore, we can maximize these quantities by minimizing the quantities in their denominators that depend on $\nu$. By considering a convex combination determined by the trade-off parameter $\kappa$ and applying scaling coefficients, we arrive at the objective in \eqref{eq:mix}.

Next, we consider the constraints in \eqref{eq:mix}. The first constraint implies $\fess(\nu) \geq Bn$ and the second constraint implies $\ftv(\nu) \geq \nicefrac{\epsilon}{2}$. Therefore, these constraints guarantee that the effective sample size and total TV distance update size are at least as large as in the on-policy case. The remaining constraints ensure that $\nu$ is a distribution.

Finally, note that $\nu_i = \nicefrac{1}{M}, \,\, i=0,\ldots,M-1$, is a feasible solution to \eqref{eq:mix} for $B \leq M \leq 2B - 1$. Therefore, \eqref{eq:mix} is a feasible optimization problem.
\end{proof}


\begin{algorithm}[t]
\caption{Generalized Policy Improvement Algorithms}\label{alg:gpi}
\KwIn{initial policy $\pi_0$; TV distance trust region parameter $\epsilon$; on-policy batch size $N=Bn$, where $n$ represents minimum batch size; trade-off parameter $\kappa$.}
\BlankLine
Calculate mixture distribution $\nu$ using \eqref{eq:mix}, and let $M$ be the number of prior policies with non-zero weighting. 
\BlankLine
Calculate generalized trust region parameter $\epsgen$ using \eqref{eq:epsgen}.
\BlankLine
\For{$k=0,1,2,\ldots$}{
	\BlankLine
        Collect $n$ samples via trajectory rollouts $\tau \sim \pi_k$.	
	\BlankLine
	Use $n$ samples from each of $\pi_{k-i}$, $i=0,\ldots,M-1$, to approximate the expectations in \eqref{eq:gen_trust_tv_one}. 
	\BlankLine	
	Update policy by approximately solving the optimization problem in \eqref{eq:gen_trust_tv_one}. Implementation varies by algorithm.
	\BlankLine
}
\end{algorithm}


In the left-hand side of \figref{fig:gpi_nu}, we see the benefits of using the optimal mixture distribution from \thmref{thm:mix} for different values of $B$. As $B$ becomes larger, on-policy algorithms require larger batch sizes and the benefit of our generalized approach increases. In the right-hand side of \figref{fig:gpi_nu}, we see that the main benefit of optimizing $\nu$ comes from determining the appropriate number of prior policies to consider, with non-uniform weights offering additional improvements. Together, \thmref{thm:epsgen} and \thmref{thm:mix} provide theoretical support for the sample reuse we consider in this work.


\section{Generalized Policy Improvement Algorithms}\label{sec:gpi}

The theory that we have developed can be used to construct generalized versions of on-policy algorithms with theoretically supported sample reuse. We refer to this class of algorithms as Generalized Policy Improvement (GPI) algorithms. The high-level framework for these algorithms is shown in \algref{alg:gpi}. By using the optimal mixture distribution from \thmref{thm:mix}, GPI algorithms only require modest increases in memory and computation compared to on-policy algorithms. In addition, due to their use of one-step trust regions, GPI algorithms only need access to the current policy and value function in order to compute updates. As a result, deep RL implementations do not require storage of any additional neural networks compared to on-policy algorithms. 

In this section, we provide details on three algorithms from this class: Generalized PPO (GePPO), Generalized TRPO (GeTRPO), and Generalized VMPO (GeVMPO). 


\subsection{Generalized PPO}

PPO approximates the on-policy trust region update in \eqref{eq:on_trust_tv} using the policy update
\begin{multline}\label{eq:ppo}
\pi_{k+1} = \arg \max_{\pi} \E_{(s,a) \sim d^{\pi_k}} \left[ \min \left( \frac{\pi(a \mid s)}{\pi_k(a \mid s)} \A, \right. \right. \\ \left. \left. \operatorname{clip} \left( \frac{\pi(a \mid s)}{\pi_k(a \mid s)},1-\epsilon,1+\epsilon \right) \A \right) \right],
\end{multline}
where $\operatorname{clip}(x,l,u) = \min ( \max (x,l), u )$ and the maximization is implemented using minibatch stochastic gradient ascent. The policy update in \eqref{eq:ppo} approximately maximizes a lower bound on the on-policy surrogate objective, while also heuristically enforcing the on-policy trust region in \eqref{eq:on_trust_tv} through the use of the clipping mechanism 
\begin{equation*}
 \operatorname{clip} \left( \frac{\pi(a \mid s)}{\pi_k(a \mid s)},1-\epsilon,1+\epsilon \right)
\end{equation*}
in the second term \cite{queeney_2021_geppo}. The clipping mechanism accomplishes this by removing any incentive for the probability ratio to deviate more than $\epsilon$ from its starting point during every policy update. 

In order to develop a generalized version of PPO that approximates the generalized trust region update in \eqref{eq:gen_trust_tv_one}, we desire a clipping mechanism that heuristically enforces the generalized trust region in \eqref{eq:gen_trust_tv_one}. As shown in \cite{queeney_2021_geppo}, the clipping mechanism 
\begin{equation*}
\operatorname{clip} \bigg( \frac{\pi(a \mid s)}{\pi_{k-i}(a \mid s)},\, \\ \frac{\pi_{k}(a \mid s)}{\pi_{k-i}(a \mid s)}-\epsgen,\, \frac{\pi_{k}(a \mid s)}{\pi_{k-i}(a \mid s)}+\epsgen \bigg)
\end{equation*}
accomplishes this goal by removing the incentive for the probability ratio to deviate more than $\epsgen$ from its starting point of $\nicefrac{\pi_{k}(a \mid s)}{\pi_{k-i}(a \mid s)}$. By applying this generalized clipping mechanism and considering a lower bound on the generalized surrogate objective, we arrive at the Generalized PPO (GePPO) update
\begin{multline}\label{eq:geppo}
\pi_{k+1} = \arg \max_{\pi} \E_{i \sim \nu} \Bigg[ \E_{(s,a) \sim d^{\pi_{k-i}}} \bigg[ \\ \min  \bigg( \frac{\pi(a \mid s)}{\pi_{k-i}(a \mid s)} \A,    \operatorname{clip} \bigg( \frac{\pi(a \mid s)}{\pi_{k-i}(a \mid s)},\, \\ \frac{\pi_{k}(a \mid s)}{\pi_{k-i}(a \mid s)}-\epsgen,\, \frac{\pi_{k}(a \mid s)}{\pi_{k-i}(a \mid s)}+\epsgen \bigg) \A \bigg) \bigg] \Bigg].
\end{multline}
See \cite{queeney_2021_geppo} for additional details, where we first introduced GePPO.


\subsection{Generalized TRPO}

TRPO approximates the on-policy update in \eqref{eq:on_trust_tv} by instead applying a forward KL divergence trust region, leading to the policy update
\begin{equation}\label{eq:trpo}
\begin{split}
\pi_{k+1} = \arg \max_{\pi} \,\, & \E_{(s,a) \sim d^{\pi_k}} \left[ \frac{\pi(a \mid s)}{\pi_k(a \mid s)} \A \right] \\
\textnormal{s.t.} \,\, & \E_{s \sim d^{\pi_k}} \left[ \KL \left( \pi_k \Vert \pi \right)(s) \right] \leq \delta,
\end{split}
\end{equation}
where $\KL \left( \pi_k \Vert \pi \right)(s)$ represents the forward KL divergence of the distribution $\pi(\, \cdot \mid s)$ from the distribution $\pi_k(\, \cdot \mid s)$. For $\delta = \nicefrac{\epsilon^2}{2}$, an application of Pinsker's inequality~\cite{tsybakov_2009} followed by Jensen's inequality shows that
\begin{equation*}
\begin{split}
\E_{s \sim d^{\pi_k}} \left[ \TV \left( \pi, \pi_k \right)(s) \right] \, &\leq \, \sqrt{\frac{1}{2} \E_{s \sim d^{\pi_k}} \left[ \KL \left( \pi_k \Vert \pi \right)(s) \right]} \\
&\leq \, \sqrt{\frac{\delta}{2}} \, = \, \frac{\epsilon}{2},
\end{split}
\end{equation*}
so the update in \eqref{eq:trpo} satisfies the trust region in \eqref{eq:on_trust_tv}.

TRPO considers a first-order expansion of the surrogate objective and second-order expansion of the forward KL divergence trust region in \eqref{eq:trpo}, leading to the approximations
\begin{gather*}
\E_{(s,a) \sim d^{\pi_k}} \left[ \frac{\pi(a \mid s)}{\pi_k(a \mid s)} \A \right] \approx \mathbf{g}_k'\left( \boldsymbol\theta - \boldsymbol\theta_k \right), \\
\E_{s \sim d^{\pi_k}} \left[ \KL \left( \pi_k \Vert \pi \right)(s) \right] \approx \frac{1}{2} \left( \boldsymbol\theta - \boldsymbol\theta_k \right)' \mathbf{F}_k \left( \boldsymbol\theta - \boldsymbol\theta_k \right),
\end{gather*}
where $\boldsymbol\theta, \boldsymbol\theta_k \in \boldsymbol\Theta$ are the parameterizations of $\pi$ and $\pi_k$, respectively. Using these approximations, the TRPO policy update admits a closed-form solution for the parameterization of $\pi_{k+1}$ given by $\boldsymbol\theta_{k+1} = \boldsymbol\theta_k + \beta \mathbf{v}$, where $\mathbf{v} = \mathbf{F}_k^{-1} \mathbf{g}_k$ is the update direction and $\beta = \sqrt{\nicefrac{2 \delta}{\mathbf{v}'\mathbf{F}_k \mathbf{v}}}$. Note that the update direction cannot be calculated directly in high dimensions, so instead it is approximated by applying a finite number of conjugate gradient steps to $\mathbf{F}_k \mathbf{v} = \mathbf{g}_k$. Finally, a backtracking line search is performed to guarantee that the trust region in \eqref{eq:trpo} is satisfied.

It is straightforward to extend TRPO to the generalized setting. We approximate the generalized update in \eqref{eq:gen_trust_tv_one} by instead applying a forward KL divergence trust region, leading to the policy update 
\begin{align}
\pi_{k+1} = \arg \max_{\pi} \,\, & \E_{i \sim \nu} \left[ \E_{(s,a) \sim d^{\pi_{k-i}}} \left[ \frac{\pi(a \mid s)}{\pi_{k-i}(a \mid s)} \A \right] \right] \nonumber \\
\textnormal{s.t.} \,\, & \E_{i \sim \nu} \left[ \E_{s \sim d^{\pi_{k-i}}} \left[ \KL \left( \pi_k \Vert \pi \right)(s) \right] \right] \leq \deltagen \label{eq:getrpo}
\end{align}
with $\deltagen = \nicefrac{\epsgen^2}{2}$. We consider the same approximations and optimization procedure as TRPO to implement the Generalized TRPO (GeTRPO) update.


\subsection{Generalized VMPO}

VMPO approximates the on-policy update in \eqref{eq:on_trust_tv} by instead considering a reverse KL divergence trust region, and treats the state-action visitation distribution as the variable rather than the policy. We write the new and current state-action visitation distributions $\psi, \psi_k$ as
\begin{equation*}
\psi(s,a) = d^{\pi_k}(s) \pi(a \mid s), \quad \psi_k(s,a) = d^{\pi_k}(s) \pi_k(a \mid s),
\end{equation*}
which results in the non-parametric VMPO update
\begin{equation}\label{eq:vmpo_np}
\begin{split}
\psi_{\textnormal{targ}} = \arg \max_{\psi} \,\, & \E_{(s,a) \sim \psi} \left[ \A \right] \\
\textnormal{s.t.} \,\, & \KL \left( \psi \Vert \psi_k \right) \leq \delta.
\end{split}
\end{equation}
As shown in \cite{song_2020}, the target distribution can be written as
\begin{equation*}
\psi_{\textnormal{targ}}(s,a) = d^{\pi_k}(s) \pi_{k}(a \mid s) w(s,a),    
\end{equation*}
where
\begin{equation*}
\begin{split}
w(s,a) &= \exp\left( \nicefrac{\A}{\lambda^*} \right) \, / \, Z(\lambda^*), \\
Z(\lambda^*) &= \E_{(s,a) \sim d^{\pi_k}} \left[ \exp \left( \nicefrac{\A}{\lambda^*} \right) \right],
\end{split}
\end{equation*}
and
\begin{equation*}
\lambda^* = \arg \min_{\lambda \geq 0} \, \, \lambda \delta + \lambda \log \left( Z(\lambda) \right)
\end{equation*}
is the optimal solution to the corresponding dual problem.

Next, VMPO projects this target distribution back onto the space of parametric policies, while guaranteeing that the new policy satisfies a forward KL divergence trust region. Therefore, VMPO guarantees approximate policy improvement in both the initial non-parametric step and the subsequent projection step. This results in the constrained maximum likelihood update
\begin{equation}\label{eq:vmpo_mle}
\begin{split}
\pi_{k+1} = \arg \max_{\pi} \,\, & \E_{(s,a) \sim d^{\pi_k}} \left[  w(s,a)  \log \pi(a \mid s) \right] \\
\textnormal{s.t.} \,\, & \E_{s \sim d^{\pi_k}} \left[ \KL \left( \pi_k \Vert \pi \right)(s) \right] \leq \delta,
\end{split}
\end{equation}
which we implement by applying the same approximation techniques used in TRPO.


\begin{figure}
\centering
\includegraphics[width=1.00\linewidth]{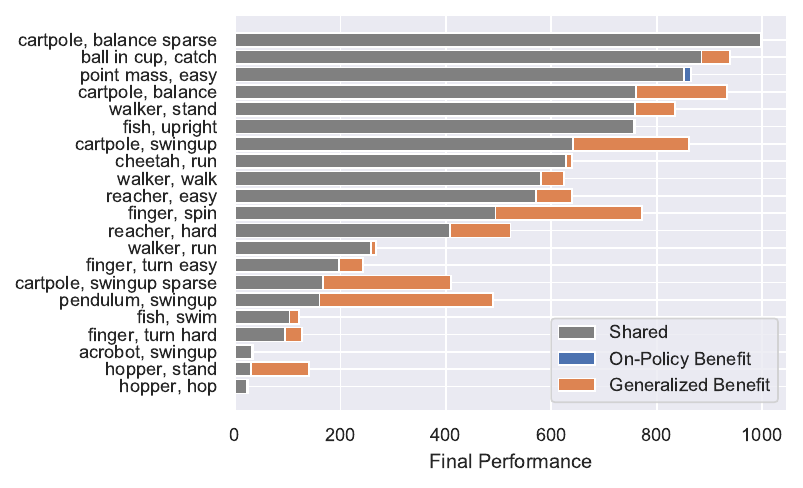}
\caption{Generalized vs.~on-policy final performance by task. Bars represent final performance of the best performing GPI algorithm and the best performing on-policy algorithm. Excludes 7 tasks where no learning occurs under any algorithm. Sorted from high to low based on on-policy performance.}\label{fig:J}
\end{figure}


In order to generalize VMPO, we begin by approximating the generalized update in \eqref{eq:gen_trust_tv_one} with a reverse KL divergence trust region, and we transform the update to consider state-action visitation distributions given by 
\begin{equation*}
\begin{split}
\psi(s,a) &= \E_{i \sim \nu} \left[ d^{\pi_{k-i}}(s) \right] \pi(a \mid s), \\
\psi_k(s,a) &= \E_{i \sim \nu} \left[ d^{\pi_{k-i}}(s) \right] \pi_{k}(a \mid s).
\end{split}
\end{equation*}
Using these generalized visitation distributions, the non-parametric update has the same form as \eqref{eq:vmpo_np} with $\delta$ replaced by $\deltagen$. This results in the target distribution
\begin{equation*}
\psi_{\textnormal{targ}}(s,a) = \E_{i \sim \nu} \left[ d^{\pi_{k-i}}(s) \right] \pi_{k}(a \mid s) w(s,a),
\end{equation*}
where $w(s,a)$ has the same form as in the on-policy case, the normalizing coefficient is now given by
\begin{equation*}
Z(\lambda^*) = \E_{i \sim \nu} \left[ \E_{(s,a) \sim d^{\pi_{k-i}}} \left[ \frac{\pi_k(a \mid s)}{\pi_{k-i}(a \mid s)} \exp \left( \nicefrac{\A}{\lambda^*} \right) \right] \right],
\end{equation*}
and $\lambda^*$ is the optimal solution to the corresponding dual problem as in the on-policy case.

The projection step in the generalized case considers a forward KL divergence trust region, resulting in the Generalized VMPO (GeVMPO) update
\begin{align}
\pi_{k+1} = \arg \max_{\pi} \,\, & \E_{i \sim \nu} \Bigg[ \E_{(s,a) \sim d^{\pi_{k-i}}} \bigg[ \nonumber \\ & \qquad  \frac{\pi_k(a \mid s)}{\pi_{k-i}(a \mid s)} w(s,a)  \log \pi(a \mid s) \bigg] \Bigg] \nonumber \\
\textnormal{s.t.} \,\, & \E_{i \sim \nu} \left[ \E_{s \sim d^{\pi_{k-i}}} \left[ \KL \left( \pi_k \Vert \pi \right)(s) \right] \right] \leq \deltagen. \label{eq:gevmpo_mle}
\end{align}


\section{Experiments}\label{sec:experiments}


\begin{figure}
\centering
\includegraphics[width=1.00\linewidth]{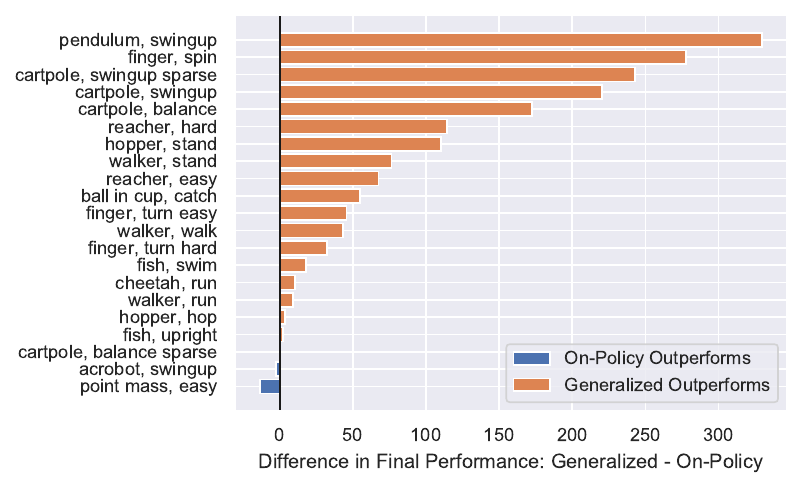}
\caption{Difference between generalized and on-policy final performance by task. Bars represent difference in final performance between the best performing GPI algorithm and the best performing on-policy algorithm. Excludes 7 tasks where no learning occurs under any algorithm. Sorted from high to low.}\label{fig:Jdiff}
\end{figure}


In order to analyze the performance of our GPI algorithms, we consider the full set of 28 continuous control benchmark tasks in the DeepMind Control Suite \cite{tunyasuvunakool_2020}. This benchmark set covers a broad range of continuous control tasks, including a variety of classic control, goal-oriented manipulation, and locomotion tasks. In addition, the benchmark tasks vary in complexity, both in terms of dimensionality and sparsity of reward signals. Finally, each task has a horizon length of $1{,}000$ and $r(s,a) \in [0,1]$ for every state-action pair, resulting in a total return between $0$ and $1{,}000$.

We focus our analysis on the comparison between GPI algorithms and their on-policy policy improvement counterparts. Note that we do not claim state-of-the-art performance, but instead we are interested in evaluating the benefits of theoretically supported sample reuse in the context of policy improvement algorithms. By doing so, we can support the use of GPI algorithms in settings where on-policy methods are currently the most viable option for data-driven control.


\subsection{Implementation Details}

In our experiments, we follow \cite{queeney_2021_geppo} by considering default network architectures and hyperparameters commonly found in the literature \cite{henderson_2018,engstrom_2020,andrychowicz_2021}. We model the policy $\pi$ as a multivariate Gaussian distribution with diagonal covariance, where the mean action for a given state is parameterized by a neural network with two hidden layers of 64 units each and tanh activations. The state-independent standard deviation of each action dimension is parameterized separately. We represent the value function using a neural network with the same structure. For each task, we train our policy for a total of 1 million steps, and we average over 5 random seeds. Note that some of the more difficult, high-dimensional tasks do not demonstrate meaningful learning under these default choices for any of the policy improvement algorithms we consider. These tasks likely require significantly longer training, different network architectures, or other algorithmic frameworks to successfully learn.

We evaluate performance across three on-policy policy improvement algorithms: PPO, TRPO, and VMPO. For these on-policy algorithms, we consider the default batch size of $N=2{,}048$, which we write as $B=2$ and $n=1{,}024$ since every task we consider has a horizon length of $1{,}000$. We consider $\epsilon=0.2$ for the TV distance trust region parameter in \eqref{eq:on_trust_tv}, which corresponds to the clipping parameter $\epsilon$ in PPO. We calculate the KL divergence trust region parameter as $\delta=\nicefrac{\epsilon^2}{2}=0.02$. We estimate advantages using Generalized Advantage Estimation (GAE) \cite{schulman_2016}.

We also consider generalized versions of each on-policy algorithm: GePPO, GeTRPO, and GeVMPO. When selecting the mixture distribution over prior policies according to \thmref{thm:mix}, we consider the trade-off parameter with the best final performance from the set of $\kappa=0.0,0.5,1.0$. For $B=2$, these choices result in using data from the prior 3 or 4 policies. The generalized trust region parameter $\epsgen$ is calculated according to \thmref{thm:epsgen}, and we set $\deltagen = \nicefrac{\epsgen^2}{2}$. As in the on-policy case, our GPI algorithms require estimates of $\A$. In order to accomplish this, we consider an off-policy variant of GAE that uses the V-trace value function estimator \cite{espeholt_2018}. See \cite{queeney_2021_geppo} for additional implementation details.\footnote{Code is available at \url{https://github.com/jqueeney/gpi}.}


\subsection{Overview of Experimental Results}

In order to evaluate the benefits of our GPI framework, we compare the best performing on-policy algorithm to the best performing GPI algorithm for every task where learning occurs. \figref{fig:J} shows the final performance of these algorithms by task, and \figref{fig:Jdiff} shows the difference in final performance by task. From these results, we see a clear performance gain from our generalized approach. Out of 21 tasks where learning occurs, our GPI algorithms outperform on-policy algorithms in 19 tasks. In the 2 tasks where on-policy algorithms outperform, the performance difference is small. On the other hand, in the tasks where our GPI algorithms outperform, we often observe a significant performance difference. We see an improvement of more than 50\% from our GPI algorithms in 4 of the tasks and an improvement of more than 10\% in 13 of the tasks. Note that we also observe similar trends when comparing any on-policy algorithm to its corresponding generalized version, as summarized in \tabref{tab:taskcount}.


\begin{table}
\caption{Task Classification by Algorithm}
\label{tab:taskcount}
  \centering
  \begin{tabular}{l R{0.08} R{0.08} R{0.08} R{0.08} }
    \toprule
    Task Classification & \multicolumn{1}{c}{PPO} & \multicolumn{1}{c}{TRPO} & \multicolumn{1}{c}{VMPO} & \multicolumn{1}{c}{Best}  \\
    \midrule
On-Policy Outperforms   & 3  \hspace*{1pt} & 1  \hspace*{1pt} & 3  \hspace*{1pt} & 2  \hspace*{1pt}  \\
Generalized Outperforms & 18 \hspace*{1pt} & 19 \hspace*{1pt} & 17 \hspace*{1pt} & 19 \hspace*{1pt}  \\
No Learning             & 7  \hspace*{1pt} & 8  \hspace*{1pt} & 8  \hspace*{1pt} & 7  \hspace*{1pt}  \\
\midrule
Total Number of Tasks   & 28 \hspace*{1pt} & 28 \hspace*{1pt} & 28 \hspace*{1pt} & 28 \hspace*{1pt}  \\
	\addlinespace
    \bottomrule
  \end{tabular}
\end{table}



\subsection{Analysis of Sparse Reward Tasks}


\begin{figure}
\centering
\includegraphics[width=1.00\linewidth]{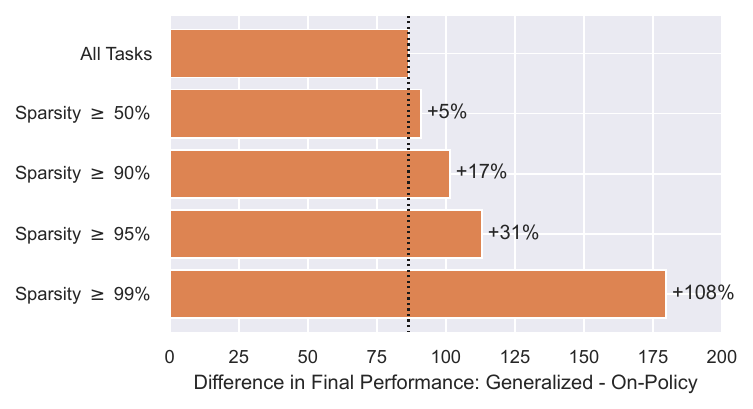}
\caption{Average difference between generalized and on-policy final performance for different sparsity levels. Bars represent difference in final performance between the best performing GPI algorithm and the best performing on-policy algorithm, averaged across all tasks with the specified sparsity level. Excludes 7 tasks where no learning occurs under any algorithm. Labels represent improvement relative to the average performance gain across all tasks, denoted by the vertical dotted line.}\label{fig:Jsparse}
\end{figure}


From \figref{fig:Jdiff}, it is clear that our GPI framework results in improved performance across a broad range of tasks. We find that this benefit is most pronounced for tasks with sparse reward signals that are difficult to find and exploit. In order to quantify the sparsity of every task in the benchmarking set, we measure the percentage of samples that contain a negligible reward signal under a random policy. We measure this statistic by collecting $100{,}000$ samples under a random policy, and we calculate the percentage of samples where $r(s,a) \leq 0.01$.

\figref{fig:Jsparse} shows the average performance gain of our GPI algorithms across tasks, where we have considered different levels of task sparsity. As we focus on tasks with increasingly sparse reward signals (i.e., a larger sparsity metric), we see that the benefits of our GPI framework also increase. For extremely sparse reward tasks where a random policy receives a reward signal less than 1\% of the time (i.e., sparsity $\geq 99\%$), the average performance gain from our GPI framework more than doubles relative to the average across all tasks. In fact, this set of extremely sparse reward tasks includes the 3 tasks where our GPI algorithms demonstrate the largest total gain in performance and the 4 tasks where our GPI algorithms demonstrate the largest percentage gain in performance. In this setting, on-policy algorithms struggle to exploit the limited reward information. The sample reuse in our GPI algorithms, on the other hand, allows sparse reward signals to be exploited across several policy updates while also leading to larger, more diverse batches of data at every update. Together, these benefits of sample reuse result in improved learning progress in difficult sparse reward settings.


\section{Conclusion}

In this work, we have developed a class of Generalized Policy Improvement algorithms that guarantee approximate policy improvement throughout training while reusing data from all recent policies. We demonstrated the theoretical benefits of principled sample reuse, and showed empirically that our generalized approach results in improved performance compared to popular on-policy algorithms across a variety of continuous control tasks from the DeepMind Control Suite. In particular, our algorithms accelerate learning in difficult sparse reward settings where on-policy algorithms perform poorly.

Because our methods use on-policy policy improvement algorithms as a starting point, our GPI algorithms only support the reuse of data from recent policies. An interesting avenue for future work includes the development of policy improvement guarantees that are compatible with the aggressive sample reuse in off-policy algorithms. Off-policy methods are often very data efficient when properly tuned, and the design of practical performance guarantees for these algorithms may increase their adoption in real-world control settings where large replay buffers are feasible.


\bibliographystyle{IEEEtran}
\bibliography{IEEEabrv,bibliography}

\begin{thebibliography}{10}
\providecommand{\url}[1]{#1}
\csname url@samestyle\endcsname
\providecommand{\newblock}{\relax}
\providecommand{\bibinfo}[2]{#2}
\providecommand{\BIBentrySTDinterwordspacing}{\spaceskip=0pt\relax}
\providecommand{\BIBentryALTinterwordstretchfactor}{4}
\providecommand{\BIBentryALTinterwordspacing}{\spaceskip=\fontdimen2\font plus
\BIBentryALTinterwordstretchfactor\fontdimen3\font minus \fontdimen4\font\relax}
\providecommand{\BIBforeignlanguage}[2]{{%
\expandafter\ifx\csname l@#1\endcsname\relax
\typeout{** WARNING: IEEEtran.bst: No hyphenation pattern has been}%
\typeout{** loaded for the language `#1'. Using the pattern for}%
\typeout{** the default language instead.}%
\else
\language=\csname l@#1\endcsname
\fi
#2}}
\providecommand{\BIBdecl}{\relax}
\BIBdecl

\bibitem{busoniu_2018}
L.~Buşoniu, T.~{de Bruin}, D.~Tolić, J.~Kober, and I.~Palunko, ``Reinforcement learning for control: Performance, stability, and deep approximators,'' \emph{Annu. Rev. Control}, vol.~46, pp. 8--28, 2018.

\bibitem{recht_2019}
B.~Recht, ``A tour of reinforcement learning: The view from continuous control,'' \emph{Annu. Rev. Control, Robot., Auton. Syst.}, vol.~2, no.~1, pp. 253--279, 2019.

\bibitem{duan_2016}
Y.~Duan, X.~Chen, R.~Houthooft, J.~Schulman, and P.~Abbeel, ``Benchmarking deep reinforcement learning for continuous control,'' in \emph{Proc. 33rd Int. Conf. Mach. Learn.}, vol.~48, 2016, pp. 1329--1338.

\bibitem{kurutach_2018}
T.~Kurutach, I.~Clavera, Y.~Duan, A.~Tamar, and P.~Abbeel, ``Model-ensemble trust-region policy optimization,'' in \emph{Proc. 6th Int. Conf. Learn. Representations}, 2018.

\bibitem{janner_2019}
M.~Janner, J.~Fu, M.~Zhang, and S.~Levine, ``When to trust your model: Model-based policy optimization,'' in \emph{Proc. Adv. Neural Inf. Process. Syst.}, vol.~32, 2019.

\bibitem{rajeswaran_2020}
A.~Rajeswaran, I.~Mordatch, and V.~Kumar, ``A game theoretic framework for model based reinforcement learning,'' in \emph{Proc. 37th Int. Conf. Mach. Learn.}, vol. 119, 2020, pp. 7953--7963.

\bibitem{wu_2019}
Y.~Wu, G.~Tucker, and O.~Nachum, ``Behavior regularized offline reinforcement learning,'' arXiv:1911.11361, 2019.

\bibitem{kumar_2020}
A.~Kumar, A.~Zhou, G.~Tucker, and S.~Levine, ``Conservative {Q}-learning for offline reinforcement learning,'' in \emph{Proc. Adv. Neural Inf. Process. Syst.}, vol.~33, 2020.

\bibitem{kidambi_2020}
R.~Kidambi, A.~Rajeswaran, P.~Netrapalli, and T.~Joachims, ``{MOReL}: Model-based offline reinforcement learning,'' in \emph{Proc. Adv. Neural Inf. Process. Syst.}, vol.~33, 2020.

\bibitem{achiam_2017}
J.~Achiam, D.~Held, A.~Tamar, and P.~Abbeel, ``Constrained policy optimization,'' in \emph{Proc. 34th Int. Conf. Mach. Learn.}, vol.~70, 2017, pp. 22--31.

\bibitem{fisac_2019}
J.~F. Fisac, A.~K. Akametalu, M.~N. Zeilinger, S.~Kaynama, J.~Gillula, and C.~J. Tomlin, ``A general safety framework for learning-based control in uncertain robotic systems,'' \emph{{IEEE} Trans. Autom. Control}, vol.~64, no.~7, pp. 2737--2752, 2019.

\bibitem{wabersich_2022}
K.~P. Wabersich, L.~Hewing, A.~Carron, and M.~N. Zeilinger, ``Probabilistic model predictive safety certification for learning-based control,'' \emph{{IEEE} Trans. Autom. Control}, vol.~67, no.~1, pp. 176--188, 2022.

\bibitem{brunke_2022}
L.~Brunke, M.~Greeff, A.~W. Hall, Z.~Yuan, S.~Zhou, J.~Panerati, and A.~P. Schoellig, ``Safe learning in robotics: From learning-based control to safe reinforcement learning,'' \emph{Annu. Rev. Control, Robot., Auton. Syst.}, vol.~5, no.~1, pp. 411--444, 2022.

\bibitem{paternain_2023}
S.~Paternain, M.~Calvo-Fullana, L.~F.~O. Chamon, and A.~Ribeiro, ``Safe policies for reinforcement learning via primal-dual methods,'' \emph{{IEEE} Trans. Autom. Control}, vol.~68, no.~3, pp. 1321--1336, 2023.

\bibitem{grossman_2023}
L.~Grossman and B.~Plancher, ``Just round: Quantized observation spaces enable memory efficient learning of dynamic locomotion,'' in \emph{IEEE Int. Conf. Robot. Automat. (ICRA)}, 2023, pp. 3002--3007.

\bibitem{queeney_2021_geppo}
J.~Queeney, I.~C. Paschalidis, and C.~G. Cassandras, ``Generalized proximal policy optimization with sample reuse,'' in \emph{Proc. Adv. Neural Inf. Process. Syst.}, vol.~34, 2021.

\bibitem{schulman_2017}
J.~Schulman, F.~Wolski, P.~Dhariwal, A.~Radford, and O.~Klimov, ``Proximal policy optimization algorithms,'' arXiv:1707.06347, 2017.

\bibitem{schulman_2015}
J.~Schulman, S.~Levine, P.~Abbeel, M.~Jordan, and P.~Moritz, ``Trust region policy optimization,'' in \emph{Proc. 32nd Int. Conf. Mach. Learn.}, vol.~37, 2015, pp. 1889--1897.

\bibitem{song_2020}
H.~F. Song, A.~Abdolmaleki, J.~T. Springenberg, A.~Clark, H.~Soyer, J.~W. Rae, S.~Noury, A.~Ahuja, S.~Liu, D.~Tirumala, N.~Heess, D.~Belov, M.~Riedmiller, and M.~M. Botvinick, ``{V-MPO}: On-policy maximum a posteriori policy optimization for discrete and continuous control,'' in \emph{Proc. 8th Int. Conf. Learn. Representations}, 2020.

\bibitem{tunyasuvunakool_2020}
S.~Tunyasuvunakool, A.~Muldal, Y.~Doron, S.~Liu, S.~Bohez, J.~Merel, T.~Erez, T.~Lillicrap, N.~Heess, and Y.~Tassa, ``dm\_control: Software and tasks for continuous control,'' \emph{Softw. Impacts}, vol.~6, p. 100022, 2020.

\bibitem{kakade_2002}
S.~Kakade and J.~Langford, ``Approximately optimal approximate reinforcement learning,'' in \emph{Proc. 19th Int. Conf. Mach. Learn.}, 2002, pp. 267--274.

\bibitem{henderson_2018}
P.~Henderson, R.~Islam, P.~Bachman, J.~Pineau, D.~Precup, and D.~Meger, ``Deep reinforcement learning that matters,'' in \emph{Proc. AAAI Conf. Artif. Intell.}, vol.~32, no.~1, 2018, pp. 3207--3214.

\bibitem{engstrom_2020}
L.~Engstrom, A.~Ilyas, S.~Santurkar, D.~Tsipras, F.~Janoos, L.~Rudolph, and A.~Madry, ``Implementation matters in deep {RL}: {A} case study on {PPO} and {TRPO},'' in \emph{Proc. 8th Int. Conf. Learn. Representations}, 2020.

\bibitem{andrychowicz_2021}
M.~Andrychowicz, A.~Raichuk, P.~Sta{\'n}czyk, M.~Orsini, S.~Girgin, R.~Marinier, L.~Hussenot, M.~Geist, O.~Pietquin, M.~Michalski, S.~Gelly, and O.~Bachem, ``What matters for on-policy deep actor-critic methods? {A} large-scale study,'' in \emph{Proc. 9th Int. Conf. Learn. Representations}, 2021.

\bibitem{queeney_2021_uatrpo}
J.~Queeney, I.~C. Paschalidis, and C.~G. Cassandras, ``Uncertainty-aware policy optimization: A robust, adaptive trust region approach,'' in \emph{Proc. AAAI Conf. Artif. Intell.}, vol.~35, no.~11, 2021, pp. 9377--9385.

\bibitem{wang_y_2019}
Y.~Wang, H.~He, X.~Tan, and Y.~Gan, ``Trust region-guided proximal policy optimization,'' in \emph{Proc. Adv. Neural Inf. Process. Syst.}, vol.~32, 2019.

\bibitem{wang_y_2020}
Y.~Wang, H.~He, and X.~Tan, ``Truly proximal policy optimization,'' in \emph{Proc. 35th Uncertainty Artif. Intell. Conf.}, vol. 115, 2020, pp. 113--122.

\bibitem{cheng_2022}
Y.~Cheng, L.~Huang, and X.~Wang, ``Authentic boundary proximal policy optimization,'' \emph{{IEEE} Trans. Cybern.}, vol.~52, no.~9, pp. 9428--9438, 2022.

\bibitem{vuong_2018}
Q.~Vuong, Y.~Zhang, and K.~Ross, ``Supervised policy update for deep reinforcement learning,'' in \emph{Proc. 7th Int. Conf. Learn. Representations}, 2019.

\bibitem{lillicrap_2016}
T.~P. Lillicrap, J.~J. Hunt, A.~Pritzel, N.~Heess, T.~Erez, Y.~Tassa, D.~Silver, and D.~Wierstra, ``Continuous control with deep reinforcement learning,'' in \emph{Proc. 4th Int. Conf. Learn. Representations}, 2016.

\bibitem{fujimoto_2018}
S.~Fujimoto, H.~van Hoof, and D.~Meger, ``Addressing function approximation error in actor-critic methods,'' in \emph{Proc. 35th Int. Conf. Mach. Learn.}, vol.~80, 2018, pp. 1587--1596.

\bibitem{haarnoja_2018}
T.~Haarnoja, A.~Zhou, P.~Abbeel, and S.~Levine, ``Soft actor-critic: {O}ff-policy maximum entropy deep reinforcement learning with a stochastic actor,'' in \emph{Proc. 35th Int. Conf. Mach. Learn.}, vol.~80, 2018, pp. 1861--1870.

\bibitem{abdolmaleki_2018}
A.~Abdolmaleki, J.~T. Springenberg, Y.~Tassa, R.~Munos, N.~Heess, and M.~Riedmiller, ``Maximum a posteriori policy optimisation,'' in \emph{Proc. 6th Int. Conf. Learn. Representations}, 2018.

\bibitem{odonoghue_2017}
B.~O'Donoghue, R.~Munos, K.~Kavukcuoglu, and V.~Mnih, ``Combining policy gradient and {Q}-learning,'' in \emph{Proc. 5th Int. Conf. Learn. Representations}, 2017.

\bibitem{gu_2017_qprop}
S.~Gu, T.~Lillicrap, Z.~Ghahramani, R.~E. Turner, and S.~Levine, ``{Q-Prop}: {S}ample-efficient policy gradient with an off-policy critic,'' in \emph{Proc. 5th Int. Conf. Learn. Representations}, 2017.

\bibitem{gu_2017_ipg}
S.~Gu, T.~Lillicrap, R.~E. Turner, Z.~Ghahramani, B.~Sch\"{o}lkopf, and S.~Levine, ``Interpolated policy gradient: {M}erging on-policy and off-policy gradient estimation for deep reinforcement learning,'' in \emph{Proc. Adv. Neural Inf. Process. Syst.}, vol.~30, 2017.

\bibitem{meng_2022}
W.~Meng, Q.~Zheng, Y.~Shi, and G.~Pan, ``An off-policy trust region policy optimization method with monotonic improvement guarantee for deep reinforcement learning,'' \emph{{IEEE} Trans. Neural Netw. Learn. Syst.}, vol.~33, no.~5, pp. 2223--2235, 2022.

\bibitem{fakoor_2020}
R.~Fakoor, P.~Chaudhari, and A.~J. Smola, ``{P3O}: {Policy-on} policy-off policy optimization,'' in \emph{Proc. 35th Uncertainty Artif. Intell. Conf.}, vol. 115, 2020, pp. 1017--1027.

\bibitem{wang_z_2017}
Z.~Wang, V.~Bapst, N.~Heess, V.~Mnih, R.~Munos, K.~Kavukcuoglu, and N.~de~Freitas, ``Sample efficient actor-critic with experience replay,'' in \emph{Proc. 5th Int. Conf. Learn. Representations}, 2017.

\bibitem{novati_2019}
G.~Novati and P.~Koumoutsakos, ``Remember and forget for experience replay,'' in \emph{Proc. 36th Int. Conf. Mach. Learn.}, vol.~97, 2019, pp. 4851--4860.

\bibitem{wang_c_2020}
C.~Wang, Y.~Wu, Q.~Vuong, and K.~Ross, ``Striving for simplicity and performance in off-policy {DRL}: Output normalization and non-uniform sampling,'' in \emph{Proc. 37th Int. Conf. Mach. Learn.}, vol. 119, 2020, pp. 10\,070--10\,080.

\bibitem{schaul_2016}
T.~Schaul, J.~Quan, I.~Antonoglou, and D.~Silver, ``Prioritized experience replay,'' in \emph{Proc. 4th Int. Conf. Learn. Representations}, 2016.

\bibitem{debruin_2018}
T.~de~Bruin, J.~Kober, K.~Tuyls, and R.~Babu{\v{s}}ka, ``Experience selection in deep reinforcement learning for control,'' \emph{J. Mach. Learn. Res.}, vol.~19, no.~9, pp. 1--56, 2018.

\bibitem{kong_1992}
A.~Kong, ``A note on importance sampling using standardized weights,'' Tech. Rep. 348, Dept. Statist., Univ. Chicago, 1992.

\bibitem{hong_2018}
Z.-W. Hong, T.-Y. Shann, S.-Y. Su, Y.-H. Chang, T.-J. Fu, and C.-Y. Lee, ``Diversity-driven exploration strategy for deep reinforcement learning,'' in \emph{Proc. Adv. Neural Inf. Process. Syst.}, vol.~31, 2018.

\bibitem{tsybakov_2009}
A.~B. Tsybakov, \emph{Introduction to Nonparametric Estimation}.\hskip 1em plus 0.5em minus 0.4em\relax Springer New York, NY, 2009.

\bibitem{schulman_2016}
J.~Schulman, P.~Moritz, S.~Levine, M.~I. Jordan, and P.~Abbeel, ``High-dimensional continuous control using generalized advantage estimation,'' in \emph{Proc. 4th Int. Conf. Learn. Representations}, 2016.

\bibitem{espeholt_2018}
L.~Espeholt, H.~Soyer, R.~Munos, K.~Simonyan, V.~Mnih, T.~Ward, Y.~Doron, V.~Firoiu, T.~Harley, I.~Dunning, S.~Legg, and K.~Kavukcuoglu, ``{IMPALA}: {S}calable distributed deep-{RL} with importance weighted actor-learner architectures,'' in \emph{Proc. 35th Int. Conf. Mach. Learn.}, vol.~80, 2018, pp. 1407--1416.

\end{thebibliography}


\end{document}